\documentclass[runningheads]{llncs}

\usepackage{eccv}
\usepackage{eccvabbrv}

\usepackage{graphicx}
\usepackage{tikz}
\usetikzlibrary{positioning,arrows,calc,fit,backgrounds}
\tikzset{
  bx/.style   ={draw,align=center,font=\scriptsize,inner sep=3pt,minimum height=6.5mm},
  dk/.style   ={draw,fill=black!82,text=white,align=center,font=\scriptsize\bfseries,inner sep=3pt,minimum height=6.5mm},
  lt/.style   ={draw,fill=black!6,align=center,font=\scriptsize,inner sep=3pt,minimum height=6.5mm},
  tool/.style ={draw,dashed,fill=black!3,align=center,font=\scriptsize,inner sep=3pt,minimum height=5.5mm},
  lbl/.style  ={font=\scriptsize\itshape,text=black!60},
  ar/.style   ={-latex,thick,black!70},
  arb/.style  ={-latex,thick,black!70,dashed},
}
\usepackage{booktabs}
\usepackage{array}
\usepackage[utf8]{inputenc}
\usepackage[colorlinks,linkcolor=black,citecolor=black,urlcolor=blue]{hyperref}

\newcommand{\ii}{\texttt{\$interrupt\$}}
\newcommand{\sil}{\texttt{\$silent\$}}

\usepackage{fontawesome5}
\definecolor{linkblue}{RGB}{37,99,235}
\hypersetup{urlcolor=linkblue}
\newcommand{\hflogo}{\raisebox{-0.28ex}{\includegraphics[height=1.0em]{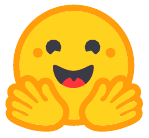}}}
\newcommand{\reslink}[4]{{#1~\textbf{#2:}~\href{#3}{#4}}}

\begin{document}

\title{Ambient @ EgoProactive 2026 : Proactive Egocentric Assistance with Visually Grounded Supervision}

\titlerunning{Speak or Stay Silent}

\author{Logesh Kumar Umapathi}
\authorrunning{L.\,K.\ Umapathi}
\institute{Team \textsc{Ambient} \\
\email{logeshkumaru@gmail.com}}

\maketitle

\begin{center}\vspace{-1mm}\small
\reslink{\faGithub}{GitHub}{https://github.com/ambient-intelligence-hq/egoproactive-verbalizer}{github.com/ambient-intelligence-hq/egoproactive-verbalizer}\\[2pt]
\reslink{\hflogo}{Hugging Face}{https://huggingface.co/collections/ambient-intelligence-labs/wearables-ai-workshop-eccv-2026}{Models \& Datasets Collection}
\end{center}

\begin{abstract}
We present our submission to the EgoProactive track of the ECCV 2026 Wearable
AI Challenge, which ranked first in the large-model division and second in the
$\leq$2B division. The task requires a wearable assistant to decide after each
eight-second segment of egocentric video whether to intervene or remain silent.

Our approach has two main components. First, we reformulate intervention timing
as single-token classification. Rather than generating either
\ii{}$\langle$utterance$\rangle$ or \sil{}, the model predicts
\texttt{yes} or \texttt{no}, and we derive the decision from the renormalised
probabilities of these two tokens. This formulation improved macro-F1 by $0.249$
and G-mean by $0.30$ over free-form generation. Second, because labelled data
were limited to the released validation set, we generated additional supervision
using a tool-calling video agent that inspects each clip and assigns intervention
timestamps. A narration-only alternative was four times larger and ten times
cheaper, but transferred worse than supervision from an unrelated real corpus,
suggesting that visual grounding is more important than annotation volume for
this task.

\keywords{Egocentric video \and Proactive assistance \and Streaming
video-language models \and Synthetic data \and Agentic annotation}
\end{abstract}

\section{Introduction}
\label{sec:metric}

A proactive wearable assistant is intended to be a real-time , context-aware assistant that 
helps the user perform a task by providing real-time guidance, feedback and assistance. Interjection 
and timing of the assistant's intervention is crucial for the success of the task. 
The EgoProactive track of the Wearable AI Workshop~\cite{wearableaiworkshop2026}
evaluates this exactly: the model receives an egocentric video of a user performing a procedural task together with their
opening query, and after each $\sim$8\,s chunk emits either \ii{} followed by a
short utterance, or \sil{}. Submissions are scored by \emph{macro-F1}, the
unweighted mean of the per-class F1 scores.

This formulation requires balancing two failure modes: intervening too often
and remaining silent when assistance is needed. Although the official metric is
macro-F1 over the two decisions, we use the geometric mean of their per-class
F1 scores for internal model selection because it assigns no credit to policies
that collapse to either extreme. Table~\ref{tab:form} illustrates this
choice in detail. Unlike prior work that frames proactive assistance as
dialogue generation~\cite{zhang2025proassist}, this track evaluates the timing
decision itself.

The organisers released 700 validation videos spanning 135 procedural tasks and
1{,}947 scored decision points, of which $54.3\%$ require intervention; the test
set remained withheld.

Our solution comprises the following components:

\begin{enumerate}
\item A \textbf{single-token verbalizer} formulation of the speak/stay-silent
decision that decouples the binary choice from utterance generation, yielding a
tunable operating point and a $+0.249$ macro-F1 ($+0.30$ G-mean) improvement over
direct generation.
\item An \textbf{agentic annotation pipeline}, built on the open-source
\textsc{Ambient} video research agent~\cite{ambient2026}, that produces visually grounded
proactive supervision, together with a controlled comparison against
narration-derived labels showing that grounding, not volume, determines transfer.
\item A \textbf{measurement protocol} for the release-set-only regime, and
evidence that in-domain evaluation inverts the ranking of the intervention that
mattered.
\item A \textbf{provably lossless vocabulary prune} bringing a 2.2132B model to
1.9977B parameters with $544/544$ identical chunk predictions, meeting the
$\leq$2B division limit at no accuracy cost.
\end{enumerate}

\section{Method}

\subsection{The verbalizer reformulation}

The task presents as conditional generation, and our first system treated it that
way, fine-tuning the model to emit the literal target string. This formulation did not work well.
The decision of \emph{whether} to speak becomes entangled with the much harder
problem of \emph{what} to say, and the gradient signal for a binary choice is
diluted across every token of the utterance. This is particularly found challenging for small scale models that we are targeting.
The failure has a clear signature: the generative model reached interrupt precision $1.000$ at recall $0.138$,
firing on 146 of 1{,}058 true positives. It found the safe corner of the loss
surface, which macro-F1 still rewards with $0.4517$ and G-mean with $0.4004$.

We instead have the model emit exactly one token, \texttt{yes} to speak or
\texttt{no} to stay silent, and read the decision from the renormalised
probability over those two logits alone:
\begin{equation}
p_{\text{interrupt}} = \operatorname{softmax}\big([\,z_{\texttt{no}},\,
z_{\texttt{yes}}\,]\big)_{1}, \qquad
\hat{y} = \mathbb{1}\!\left[\,p_{\text{interrupt}} \geq \tau\,\right].
\end{equation}

\begin{figure}[t]
\centering
\begin{tikzpicture}[node distance=4mm]
\node[lt,minimum width=11mm] (c1) {$c_1$};
\node[lt,minimum width=11mm,right=1.2mm of c1] (c2) {$c_2$};
\node[bx,fill=black!14,minimum width=11mm,right=1.2mm of c2] (c3) {$c_3$};
\node[lbl,above=2mm of c1.north west,anchor=west] {8\,s chunks};
\node[bx,below=7mm of c1.south west,anchor=north west,text width=27mm]
     (inp) {32 cumulative frames\\[-1pt] $+$ query $+$ last 4 turns};
\node[dk,right=8mm of inp,text width=19mm] (mdl) {Qwen3.5\\[-1pt] $+$ LoRA r32};
\node[bx,right=8mm of mdl,minimum width=13mm] (yes) {\texttt{yes}};
\node[bx,below=1.2mm of yes,minimum width=13mm] (no) {\texttt{no}};
\node[lbl,above=0.8mm of yes] {2 logits};
\node[bx,right=8mm of yes,minimum width=20mm] (int) {\ii{}};
\node[bx,below=1.2mm of int,minimum width=20mm] (sil) {\sil{}};
\draw[ar] (c3.south) -- ++(0,-3mm) -| (inp.north);
\draw[ar] (inp) -- (mdl);
\draw[ar] (mdl.east) -- ($(yes.west)!0.5!(no.west)$);
\draw[ar] ($(yes.east)!0.5!(no.east)$) -- node[above,lbl,pos=0.45]{$\tau$} ($(int.west)!0.5!(sil.west)$);
\draw[arb] (sil.south) -- ++(0,-4mm) -| node[below,lbl,pos=0.25]{appended to history} (inp.south);
\end{tikzpicture}
\caption{One decision per chunk. Frames accumulate from the start of the video and
are strided to a cap of 32; the model emits a single token, and the renormalised
probability over the \texttt{yes}/\texttt{no} logits is thresholded at $\tau$. The
emitted decision re-enters the history for the next chunk, which is what prevents
the always-interrupt failure of Table~\ref{tab:form}.}
\label{fig:loop}
\end{figure}
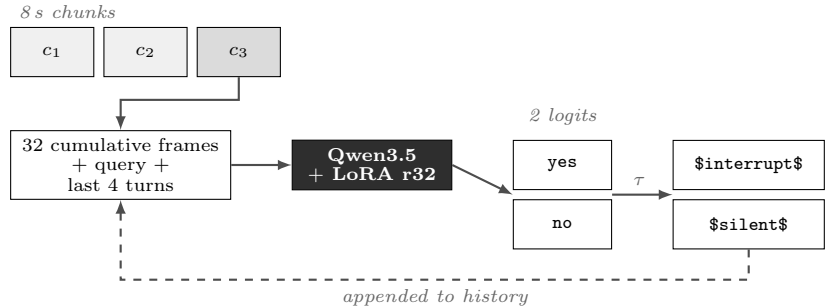

Loss is cross-entropy over the supervised span only: labels are masked
everywhere except the final assistant turn's \texttt{yes}/\texttt{no} token and
its end-of-turn marker. The utterance is discarded at training time and
templated at inference, because the metric never scores its content.

Figure~\ref{fig:loop} shows the resulting per-chunk loop. This buys three
properties at once. The gradient signal is a clean binary. The
operating point $\tau$ becomes tunable \emph{after} training, without
retraining. And a decision costs one forward pass with no decoding loop, which
matters under the evaluation harness's per-turn time budget.

\begin{table}[t]
\centering
\caption{Formulation comparison on the in-domain held-out split. Macro-F1, the
official metric, awards roughly a third of the achievable score to policies that
never fire or always fire; G-mean assigns both exactly zero. The two agree once a
model is non-degenerate.}
\label{tab:form}
\small
\begin{tabular}{lcccl}
\toprule
Formulation & Macro-F1 & G-mean & Int.\ P/R & Behaviour \\
\midrule
Base, no fine-tuning       & 0.3135 & 0.0000 & 0.00/0.00 & Always silent \\
Generative \ii{}/\sil{}     & 0.4517 & 0.4004 & 1.00/0.14 & Fires 146 of 1{,}058 \\
Verbalizer, no history     & 0.3521 & 0.0000 & 0.54/1.00 & Always interrupts \\
\textbf{Verbalizer (ours)}  & \textbf{0.7008} & \textbf{0.7007} & 0.72/--- & Calibrated \\
\bottomrule
\end{tabular}
\end{table}

\subsection{Input construction}

Each decision uses all frames observed up to that point, uniformly subsampled
once to at most 32 frames and resized to a maximum side length of 512\,px.
Applying a second subsampling step at inference time reduced temporal coverage
and substantially degraded performance.

The prompt includes the user's query and the four most recent dialogue turns,
including the assistant's previous decisions. This history is necessary to avoid
repeated interventions: without it, the model cannot determine whether
assistance has already been provided and predicts \ii{} for every chunk
(Table~\ref{tab:form}).

\subsection{Agent-generated supervision}

To augment the limited labelled data, we ran two
independent routes to generate labels.

\begin{figure}[t]
\centering
\resizebox{\textwidth}{!}{%
\begin{tikzpicture}[node distance=4mm]
\node[lt,text width=15mm] (clip) {raw clip\\[-1pt]($\sim$465\,s)};
\node[dk,right=7mm of clip,text width=20mm] (orch) {orchestrator\\[-1pt]\mdseries deepseek-v4-flash};
\node[tool,above right=3mm and 8mm of orch,text width=27mm] (t1) {\scriptsize\texttt{get\_video\_}\\[-2pt]\texttt{description}};
\node[tool,right=8mm of orch,text width=25mm]               (t2) {\texttt{search\_clip}};
\node[tool,below right=3mm and 8mm of orch,text width=25mm] (t3) {\texttt{focus\_clip}};
\node[dk,right=7mm of t2,text width=17mm] (vis) {Qwen3.6-27B\\[-1pt]\mdseries};
\begin{scope}[on background layer]
\node[draw,dotted,thick,fit=(orch)(t1)(t3)(vis),inner sep=3.2mm] (agentbox) {};
\end{scope}
\node[font=\scriptsize,text=black!60,fill=white,inner sep=1pt,above=1.6mm of agentbox.north west,anchor=west] {\textsc{Ambient} agent \ ($\leq$12 inspection windows per clip)};
\node[bx,below=17mm of orch.south west,anchor=north west,text width=32mm] (walk)
     {\textbf{walkthrough}\\[-1pt] events: $t$, utterance,\\[-1pt] visible context, rationale\\[-1pt] $+$ silent intervals};
\node[bx,right=7mm of walk,text width=27mm] (conv)
     {$-0.5$\,s onset shift\\[-1pt] $\rightarrow$ 8\,s binning};
\node[bx,right=7mm of conv,text width=25mm] (rows)
     {13{,}730 rows\\[-1pt] 53\% interrupt};
\draw[ar] (clip) -- (orch);
\draw[ar] (orch) -- (t1);  \draw[ar] (orch) -- (t2);  \draw[ar] (orch) -- (t3);
\draw[ar] (t1) -- (vis);   \draw[ar] (t2) -- (vis);   \draw[ar] (t3) -- (vis);
\draw[ar] (agentbox.south -| orch) -- (walk.north);
\draw[ar] (walk) -- (conv);
\draw[ar] (conv) -- (rows);
\end{tikzpicture}}
\caption{The annotation pipeline. The orchestrator reasons over the timeline and
issues tool calls; every deciding detail is confirmed against the vision model
before it becomes a label, so each cue is grounded in evidence the student model
can also see. Output is a structured list of events, including the intervals where the
agent deliberately chose silence, that is then shifted and binned into
per-chunk decisions.}
\label{fig:pipeline}
\end{figure}
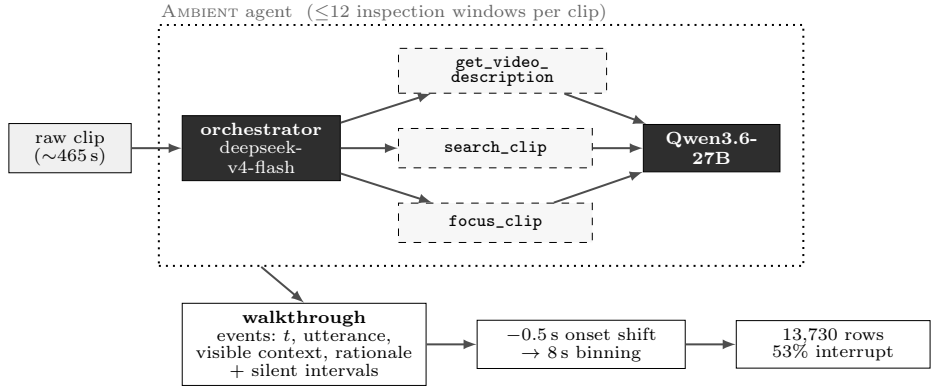

\paragraph{The agentic route.} A tool-calling agent, \textsc{Ambient}~\cite{ambient2026},
annotates each clip (Fig.~\ref{fig:pipeline}). A
\texttt{DeepSeek-V4-Flash-0731}~\cite{deepseekv4} orchestrator reasons over the timeline while a
\texttt{Qwen3.6-27B}~\cite{qwen36} vision model answers visual queries. The agent has three tools: a whole-video description
built from sparsely sampled frames, which yields a coarse map of the procedure;
and two window-level tools, one answering a question about a $\leq$5-minute
segment and one describing it in detail. The overview is explicitly treated as
approximate, so any detail that becomes a label is confirmed with a window-level
call first. Dense annotation used up to twelve inspection windows per clip to
cover a full timeline.

Output is a structured list of events, each carrying a
timestamp, a decision type, the utterance, the visible context that justified it,
a timing rationale and a confidence; plus explicit \emph{silent intervals} with
reasons. Requiring the agent to commit to where it deliberately said nothing
converts an absence of labels into a positive one. Events are then binned into
8\,s chunks with a $-0.5$\,s onset shift.

\paragraph{The narration route.} For comparison we synthesised labels from dense
human narration text: an LLM reads the narration for a clip and writes a
per-chunk speak/silent script, with no vision pass at all. This is dramatically
cheaper ($\sim$\$0.005 versus $\sim$\$0.05 per clip) and we scaled it much
further , 953 clips and 59{,}359 chunks against the agent's 234 clips and
13{,}730 rows.

\paragraph{Grounding, not volume, decides transfer.} The narration route underperformed.
A 2B model trained only on it reached $0.49$ transfer to the held-out
validation set, \emph{below} the $0.59$ obtained from an unrelated real corpus
(HoloAssist~\cite{wang2023holoassist}), with an in-domain score of only $0.52$. The diagnosis is:
narration-derived labels are weakly visually grounded. The teacher decides from
rich text while the student must predict from sparse frames at $0.83$\,fps, so
the teacher knows things the student cannot see and the student learns to guess.
The agent's labels are placed \emph{because it looked at that part of the
timeline}, so every label is reachable from the evidence the student receives.
Four times the clips did not compensate.

\subsection{The annotation policy, and how it was tuned}

The utility of the synthetic supervision depends critically on where the agent
places intervention cues. We therefore evaluated three versions of the policy
instructions included in the annotation prompt. For each version, we mapped the
agent's timestamped interventions to the challenge's official decision intervals,
pooled the resulting predictions across videos, and computed the geometric mean
of the interrupt and silent F1 scores. Policies were developed on six videos and
evaluated on six disjoint videos to detect prompt overfitting.

Error analysis of the unconstrained agent revealed three recurring failure
modes. First, it often omitted the setup phase: reference annotations commonly
begin with a materials or navigation cue near $t=0$, whereas the agent waited
until the main manipulation became visible. Second, its cues frequently lagged
the annotated action onset and fell into the following decision interval,
producing both a false negative in the intended interval and a false positive in
the adjacent one. Allowing a $\pm1$\,s matching tolerance increased G-mean by
approximately $0.05$. Third, the agent generated redundant cues during repeated
motions such as drawing, rolling, and wiping; in one example, it produced 20
cues for a task comprising roughly five semantic steps.

The final policy therefore instructed the agent to cover relevant setup actions,
place cues at action onset, merge repeated motions into a single intervention
followed by silence, and emit at most one cue per 8\,s interval. In contrast to
t2, it did not prescribe a target number or density of interventions.

\begin{table}[t]
\centering
\caption{Delivered corpus. The dense policy was never told the validation set's
interrupt frequency.}
\label{tab:corpus}
\begin{tabular}{lcl}
\toprule
Measure & Value & Note \\
\midrule
Clips annotated / submitted   & 234 / 235 & 99.6\% yield; one failure dropped \\
Training rows                 & 13{,}730  & 8\,s chunk decisions \\
Interrupt rate, dense policy  & 53\%      & against the real set's 54.3\% \\
Interrupt rate, sparse policy & 11\%      & same machinery, looser policy \\
Degenerate clips              & 0         & no all-silent or all-interrupt output \\
Cost per clip                 & $\sim$\$0.05 & self-hosted vision \\
Trajectories retained         & 10{,}081  & every tool call, auditable \\
\bottomrule
\end{tabular}
\end{table}

The base-rate agreement in Table~\ref{tab:corpus} is worth dwelling on. The dense
policy reproduced the validation set's interrupt frequency to within $1.3$ points
without being given it, by reasoning about where a coach would speak. The sparse
variant, run through identical machinery with a looser policy, landed at $11\%$.
The match is attributable to the policy, not to luck, which is why the policy
justified three iterations and a held-out check.

\subsection{Training}

Both divisions use the same recipe on different backbones (Qwen3.5-4B and
Qwen3.5-2B~\cite{qwen35}): LoRA~\cite{hu2021lora} of rank 32 and $\alpha=64$, dropout $0.05$, applied to all
seven attention and MLP projections; learning rate $1\!\times\!10^{-4}$ with a
cosine schedule and $0.03$ warmup ratio; batch size 1 with 8-step gradient
accumulation; bf16 with gradient checkpointing. The training mix is the released
videos seen twice plus the full agent-generated corpus.

\subsection{Vocabulary pruning for the $\leq$2B division}

Qwen3.5-2B~\cite{qwen35} is 2.2132B parameters once the vision tower is counted, over the
division limit. The embedding and output layers
dominate, carrying a 248{,}320-token vocabulary of which most is multilingual
coverage irrelevant to English procedural narration.

We prune contiguously, and chose this over frequency-based selection precisely
because it is unexciting. Token IDs $[0, 143000)$ are kept unchanged, and an
explicit set of higher IDs including the vision, video and special tokens the
architecture depends on is appended. Every surviving common token therefore
keeps its original index, so no re-indexing error is possible on the hot path.
Embedding rows are selected and cloned exactly, not re-initialised or projected.
Any pruned token decomposes into UTF-8 byte tokens, so no input is
unrepresentable: degraded at worst, never a crash.

\begin{table}[t]
\centering
\caption{Vocabulary pruning, verified on the 40-video container evaluation and
across all 700 released videos.}
\label{tab:prune}
\begin{tabular}{lcc}
\toprule
Property & Before & After \\
\midrule
Parameters (incl.\ vision tower) & 2.2132\,B & \textbf{1.9977\,B} \\
Vocabulary size                  & 248{,}320 & 143{,}084 \\
G-mean                           & 0.7291    & 0.7291 \\
Chunk predictions identical      & ---       & \textbf{544 / 544} \\
Agreement on $p_{\text{interrupt}}$ & ---    & $0.00\mathrm{e}{+}00$ \\
Byte-fallback incidence, 700 videos & ---    & zero \\
\bottomrule
\end{tabular}
\end{table}

The compliance constraint was met at no accuracy cost (Table~\ref{tab:prune}).

\section{Measurement}

For the final submission, we trained on part of the released validation set.
Because this reduced the amount of labelled data available for unbiased
evaluation, performance measured on the full released set could no longer
reliably indicate whether a change improved generalisation.

We used two evaluation safeguards. First, during development, we fixed a split
of the 700 released videos into 210 development and 490 held-out videos,
stratified by domain and interrupt-rate tercile. Videos used for prompt tuning
were assigned exclusively to the development partition. Second, we evaluated
candidate models on a cross-domain benchmark of 140
HoloAssist~\cite{wang2023holoassist} videos that were excluded from all
task-specific training. We ranked candidate submissions by G-mean on this
cross-domain benchmark.

\section{Results}

\begin{table}[t]
\centering
\caption{Final submissions on the cross-domain benchmark, against our best
release-set-trained models on the same 140 videos.}
\label{tab:main}
\begin{tabular}{lccc}
\toprule
Model & Cross-domain G-mean & Previous best & $\Delta$ \\
\midrule
4B, large division              & \textbf{0.5902} & 0.490 & $+0.100$ \\
2B, $\leq$2B division (pruned)  & \textbf{0.579}  & 0.476 & $+0.103$ \\
\bottomrule
\end{tabular}
\end{table}

Table~\ref{tab:leaderboard} gives the official final
standings~\cite{wearableaileaderboard2026}. Our 4.54B entry
took first place in the large division at $0.7179$ macro-F1, and our pruned
$1.9977$B entry placed second in the $\leq$2B division at $0.6866$, $0.0045$
behind the winner. These are macro-F1 on the hidden test set, and are therefore
not directly comparable with the G-mean figures used for selection elsewhere in
this report (Section~\ref{sec:metric}).

The large-division result is worth reading alongside
Section~\ref{sec:negative}: the top entry is a 4.54B model, and it is followed by
a 27B entry at $0.7055$ and a 28.9B entry at $0.5452$. Across independent teams,
a $6\times$ parameter advantage did not translate into a better score. This is
the same conclusion our internal scale study reached (Table~\ref{tab:scale}), and
it holds across differing methods.

Both entries are the same recipe at two scales (Table~\ref{tab:main}), and both
were selected on the cross-domain benchmark alone.

\begin{table}[t]
\centering
\caption{Official final leaderboard~\cite{wearableaileaderboard2026}, EgoProactive,
reported as macro-F1 on the held-out test set. Our entries are in bold. The large
division is won by a 4.54B model over 27B and 28.9B entries.}
\label{tab:leaderboard}
\small
\begin{tabular}{clcr@{\hspace{2em}}clcr}
\toprule
\multicolumn{4}{c}{\textbf{Large division (2B+)}} & \multicolumn{3}{c}{\textbf{Small division ($\leq$2B)}} \\
\cmidrule(r){1-4}\cmidrule(l){5-7}
\# & Team & Params & Macro-F1 & \# & Team & Macro-F1 \\
\midrule
\textbf{1} & \textbf{ambient (ours)} & \textbf{4.54\,B} & \textbf{0.7179} & 1 & hippo & 0.6911 \\
2 & fufu       & 4.54\,B  & 0.7127 & \textbf{2} & \textbf{ambient (ours)} & \textbf{0.6866} \\
3 & zoeyagent  & 27\,B    & 0.7055 & 3 & fufu          & 0.6677 \\
4 & teamyj     & 8.78\,B  & 0.6280 & 4 & genesis       & 0.6265 \\
5 & luludawang & 28.9\,B  & 0.5452 & 5 & sololevelling & 0.6238 \\
6 & tonytie317 & 27.8\,B  & 0.4568 & 6 & icanfly       & 0.6080 \\
7 & jt-proagent& 27\,B    & 0.4492 & 7 & shadow        & 0.5897 \\
\bottomrule
\end{tabular}
\end{table}

\begin{table}[t]
\centering
\caption{The synthetic-data ablation at equal compute. The intervention that
carried the result appears as a clear loss on the in-domain benchmark.}
\label{tab:synth}
\begin{tabular}{llccc}
\toprule
Experiment & Benchmark & Baseline & Result & $\Delta$ \\
\midrule
Agent corpus        & Cross-domain & 0.557 & 0.577 & $+0.021$ \\
Agent corpus        & In-domain    & 0.659 & 0.617 & $-0.042$ \\
+ second corpus     & Cross-domain & 0.590 & 0.548 & $-0.042$ \\
\bottomrule
\end{tabular}
\end{table}

The corpus lowers the in-domain score by $0.042$ while raising the cross-domain
score by $0.021$ (Table~\ref{tab:synth}), and the gain is concentrated in
interrupt recall ($+0.064$ F1). It teaches the model to fire on footage it has
never seen, trading fit to the released videos for robustness elsewhere. Against
a hidden test set that is the correct trade, and only the cross-domain benchmark
could see it. We read the in-domain drop not as a cost to tolerate but as
evidence that the model had stopped over-fitting the only labelled data it had.

\section{Negative results}
\label{sec:negative}

We report the failed intervention attempts because the failures were more informative than the win.

\paragraph{Changing the synthetic-data mixture did not improve transfer.}
At fixed training compute and dataset size, replacing 6.6k of the 13.7k
procedural examples with agent-annotated household and sightseeing footage
reduced cross-domain G-mean from $0.590$ to $0.548$
(Table~\ref{tab:synth}). Thus, increasing source diversity at the expense of
procedural examples was detrimental in this setting. Synthetic data derived
from Ego4D~\cite{grauman2022ego4d}, HoloAssist~\cite{wang2023holoassist}, and
Ego-Exo4D~\cite{grauman2024egoexo4d}, as well as temporal augmentation, also
produced no consistent improvement.

\paragraph{Increasing backbone size provided little benefit.}
Using the same verbalizer training recipe, the 27B model scored $0.002$ below
the 4B model, while the 2B model remained within $0.01$ of it
(Table~\ref{tab:scale}). These results suggest that, under our training setup,
increasing model capacity was less effective than improving supervision and
data composition. Consistent with this observation, our pruned 2B submission
finished within $0.005$ macro-F1 of the winner of the $\leq$2B division.

\begin{table}[t]
\centering
\caption{Backbone scale, in-domain held-out, verbalizer recipe throughout.}
\label{tab:scale}
\begin{tabular}{lccc}
\toprule
Backbone & Parameters & LoRA rank & G-mean \\
\midrule
Qwen3.6-27B & 27\,B & 32 & 0.6989 \\
Qwen3.5-4B  & 4\,B  & 32 & \textbf{0.7007} \\
Qwen3.5-2B  & 2\,B  & 64 & 0.6914 \\
\bottomrule
\end{tabular}
\end{table}

\section{Conclusion}

Two design choices contributed most strongly to our results. First, replacing
free-form generation with single-token classification improved G-mean by
approximately $0.30$, indicating that separating the intervention decision from
response generation substantially simplified the task. Second, supervision
generated by a visually grounded agent transferred better than a larger and
cheaper corpus derived from narration alone, whose labels were often based on
information unavailable in the student's visual input.

Our broader finding concerns model selection under limited labelled data.
Performance on the released in-domain set was not a reliable proxy for
cross-domain transfer: some models achieved high in-domain but only
$0.49$ cross-domain, while adding the agent-generated corpus reduced in-domain
G-mean by $0.042$ despite improving cross-domain performance. These observations
motivated selecting models on a separate cross-domain benchmark. In future work,
we would establish such an evaluation set before training and use it alongside
in-domain results to distinguish improved transfer from increased fit to the
released data.

\section*{Acknowledgments}
We thank the challenge organisers for the benchmark and the evaluation
infrastructure. The author has no competing interests to declare.

\bibliographystyle{splncs04}
\bibliography{refs}

\end{document}